# An interpretable imbalanced semi-supervised deep learning framework for improving differential diagnosis of skin diseases

Futian Weng[a,b,c,1], Yuanting Ma[d,1], Jinghan Sun[a,b], Shijun Shan[e], Qiyuan Li[f,b], Jianping Zhu[b,g,c,*], Yang Wang[h,*] and Yan Xu[i,j,*]

[a]School of Medicine, Xiamen University, Xiamen, 361005, China
[b]National Institute for Data Science in Health and Medicine, Xiamen University, Xiamen, 361005, China
[c]Data Mining Research Center, Xiamen University, Xiamen, 361005, China
[d]School of Economics and Management, East China Jiaotong University, Nanchang, 330013, China
[e]Department of Dermatology and Venereology, Xiang'an Hospital of Xiamen University, School of Medicine, Xiamen University, Xiamen, 361101, China
[f]Department of Pediatrics, the First Affiliated Hospital of Xiamen University, Xiamen, 361003, China
[g]School of Management, Xiamen University, Xiamen, 361005, China
[h]School of Science, Hong Kong University of science and technology, Hong Kong, 00852, China
[i]School of Mathematical Sciences, Ocean University of China, Qingdao 266100, China
[j]National Economic Engineering Laboratory, Dongbei University of Finance and Economics, Dalian 116025, China

ABSTRACT

Dermatological diseases are among the most common disorders worldwide. This paper presents the first study of the interpretability and imbalanced semi-supervised learning of the multiclass intelligent skin diagnosis framework (ISDL) using 58,457 skin images with 10,857 unlabelled samples. Pseudo-labelled samples from minority classes have a higher probability at each iteration of class-rebalancing self-training, thereby promoting the utilisation of unlabelled samples to solve the class imbalance problem. Our ISDL achieved a promising performance with an accuracy of 0.979, sensitivity of 0.975, specificity of 0.973, macro-F1 score of 0.974 and area under the receiver operating characteristic curve (AUC) of 0.999 for multi-label skin disease classification. The SHapleyAdditive exPlanation (SHAP) method is combined with our ISDL to explain how the deep learning model makes predictions. This finding is consistent with the clinical diagnosis. We also proposed a sampling distribution optimisation strategy to select pseudo-labelled samples in a more effective manner using ISDLplus. Furthermore, it has the potential to relieve the pressure placed on professional doctors, as well as help with practical issues associated with a shortage of such doctors in rural areas.

## 1. Introduction

Dermatosis is one of the most common human diseases. According to the National Health Commission of the People's Republic of China, despite there being fewer than 30,000 registered dermatologists, the total number of dermatological outpatients in China is 300 million per year (Commission et al., 2020). Seventy percent of skin disorders can be treated in grassroots medical institutions. However, the personnel in such institutions in China still have limited expertise in skin disease diagnosis and treatment. Individual differences in the focus shape, body position distribution, colour, size, and arrangement make accurate skin disease diagnosis difficult. Usually, only experienced doctors can obtain a high diagnostic accuracy using clinical diagnostic methods (Saez et al., 2014; Habif et al., 2017). Furthermore, human specialists have limitations in terms of the diagnosis, which is primarily reliant on subjective assessment, and various expert judgments differ substantially (Madooei and Drew, 2016). Although many automated dermatological disease diagnosis methods have been developed, a complete decision support system has yet to be developed (Liu et al., 2020; Li et al., 2020).

With the rapid expansion of information resources, advancements in hardware capabilities, and big data technology, an increasing number of studies have incorporated artificial intelligence technology into auxiliary dermatology diagnoses (Goceri, 2019; Iqbal, 2021; Hameed et al., 2021; Janoria et al., 2021). Many studies have shown that deep learning methods can surpass human beings in many computer vision tasks such as image classification (Chan et al., 2015; Korot et al., 2021), semantic segmentation (Garcia-Garcia et al., 2017; Ouahabi and Taleb-Ahmed, 2021), and object detection and location (Zhao et al., 2019; Wang et al., 2021a). A key factor in this success is the ability of certain neural network algorithms to capture nonlinear representations from complex high-dimensional data (Bengio et al., 2013; LECUN, 2015). In particular, such algorithms have been used for the diagnosis of skin diseases. For example, Wang et al. developed a deep learning framework for skin lesion segmentation and melanoma recognition and achieved the highest performance in comparison with other methods (Wang et al., 2021b). In addition, Liu et al. (2020) provided a differential diagnosis of skin conditions using deep learning models, resulting in a superior performance that was equal to that of six other dermatologists and better than that of six primary care physicians. These findings suggest that deep learning has become an effective method for di-

*Corresponding author.E-mail address: xmjpzhu@163.com (J.P.Zhu), yangwang@ust.hk (Y.Wang), yan_xu@dufe.edu.cn (Y.Xu)

[1]These authors contributed equally.

agnosing skin diseases and has attracted extensive attention. The performance of Dermatology's intelligent diagnosis approach based on deep learning is comparable to that of actual dermatologists.

Despite these advancements, an intelligence-aided diagnosis of skin diseases is still in its infancy, and there are still limitations to be resolved: 1) The paucity of clinical data with a high number of markers prevents in-depth learning from being widely used in dermatology diagnosis. Although public dermatological datasets and websites can provide a large number of unlabelled data, the labelling of a large number of dermatology data requires professional knowledge, which is challenging and costly. 2) The class imbalance problem is widespread in the intelligent diagnosis of skin diseases, making the prediction results of the model more inclined to major categories. Learning a robust and high-accuracy model is a challenge. 3) To a certain extent, the high-accuracy prediction of a deep learning method depends on the complexity of the network model. Therefore, a deep learning model usually has many parameters and is inefficient in terms of training and application. 4) The use of a "black box" has always been a controversial topic in the application of deep learning models, and its opacity is a bottleneck.

In this study, an interpretable semi-supervised deep learning multi-category intelligent skin disease diagnosis system is established. The framework was built on a deep learning model with a few parameters and high accuracy. It can realise an accurate and rapid intelligent diagnosis of skin disease images. Minor samples with higher probability are selected for the pseudo-label dataset, improving the model prediction accuracy and reducing the amount of image annotation required. Finally, the SHAP approach was used to locally interpret the deep learning model, quantify the contribution of pixels in each skin disease image to the final prediction, and provide the reason for the prediction from the deep learning model.

## 2. Materials and methods

### 2.1. Dataset

We collected a dataset from the International Skin Imaging Collaboration (ISIC), which aggregates a large-scale publicly available dataset of skin disease images (Codella et al., 2019). The dataset contained 58,457 images, 10,857 of which are unlabelled. Finally, nine categories were obtained by mapping the labels of all data: actinic keratosis (AK), basal-cell carcinoma (BCC), benign lichenoid keratosis (BKL), dermatofibroma (DF), melanoma (MEL), naevus (NV), squamous cell carcinoma (SCC), vascular disease (VASC), and unknown (UK) (see Table 1).

The sample distribution of these categories is shown in Table 2, and the data distribution shows the characteristics of a class imbalance. The number of majority classes divided by the minority classes was 113.5. Based on the above, we split the image data into several sets and defined the following: Suppose that the labelled dataset is $X = \{(x_n, y_n) : n \in (1, \cdots, N)\}$ where $L$-class and $x_n$ are training instances, and $y_n \in \{1, \cdots, L\}$ denote the corresponding labels. We denote the number of training instances in $X$ of class $l$ as $N_l$, i.e., $\sum_{l=1}^{L} N_l = N$. The unlabelled samples are denoted by $Z = \{z_m \in \mathbb{R}^d : m \in (1, \cdots, M)\}$.

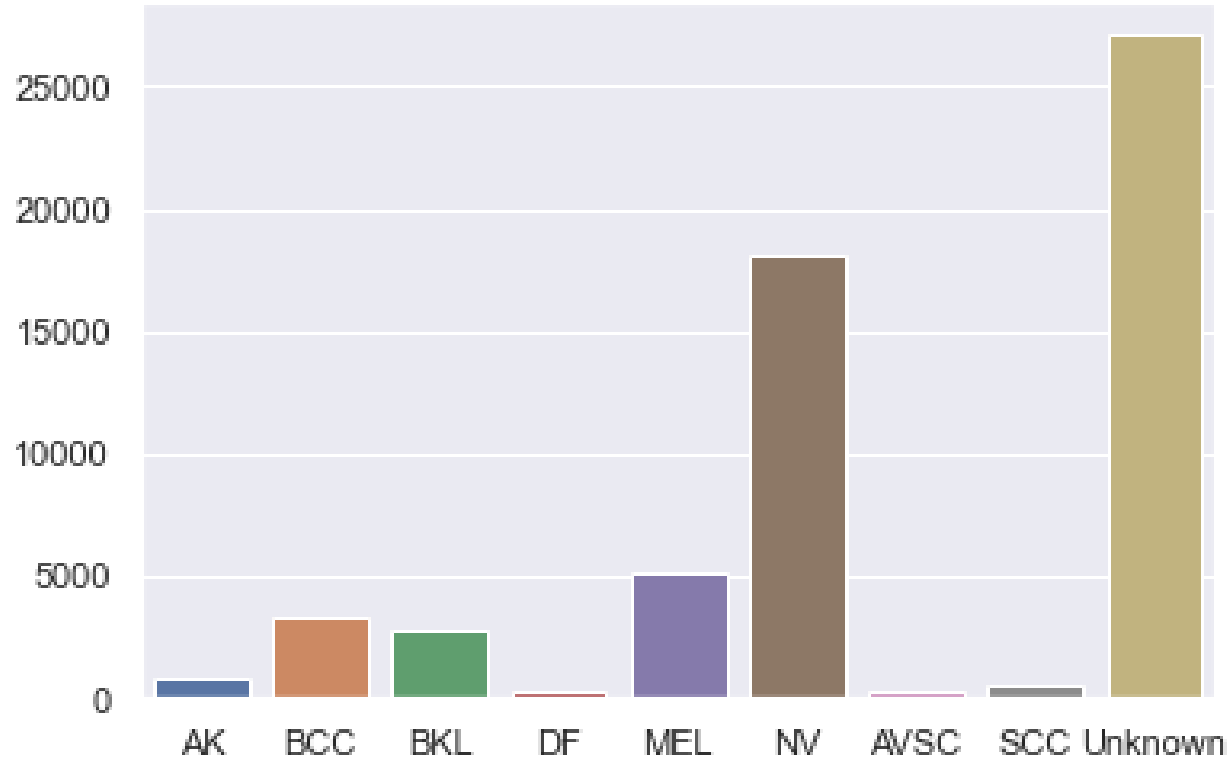

**Figure 1:** Illustration of class distribution

**Table 1**
Mapping from diagnosis to targets

| 2018 | 2019 | 2020 | Target |
|---|---|---|---|
| nv | NV | naevusn | NV |
| mel | MEL | Melanoma | MEL |
| bcc | BCC | | BCC |
| bkl | BKL | NOS[a] | BKL |
| akiec | AK | | AK |
| | SCC | | SCC |
| vasc | VASC | | VASC |
| df | DF | | DF |
| | | Atypical[b] | Unknown |

[a] denotes Seborrheic keratosis lichenoid keratosis Solar lentigo lentigo NOS.
[b] denotes cafe-au-lait macule atypical melanocytic proliferation unknown.

### 2.2. An overview of our proposed framework

The research framework for the intelligent diagnosis model for skin diseases proposed in this study is illustrated in Figure 2, including the data input module, construction and training of the teacher model, retraining of unlabelled samples, and the interpretable part of the model. All skin image data were split into two parts, labelled and unlabelled datasets, and input into the model. After the image augmentation, the labelled dataset was used to train a supervised skin disease classification network, called the teacher model. This model was then used to predict the unlabelled samples and generate pseudo-labels in each iteration. The optimal classification model was obtained through a cross validation. Subsequently, the SHAP method is introduced to explain the deep learning classifier model in a visualised form.

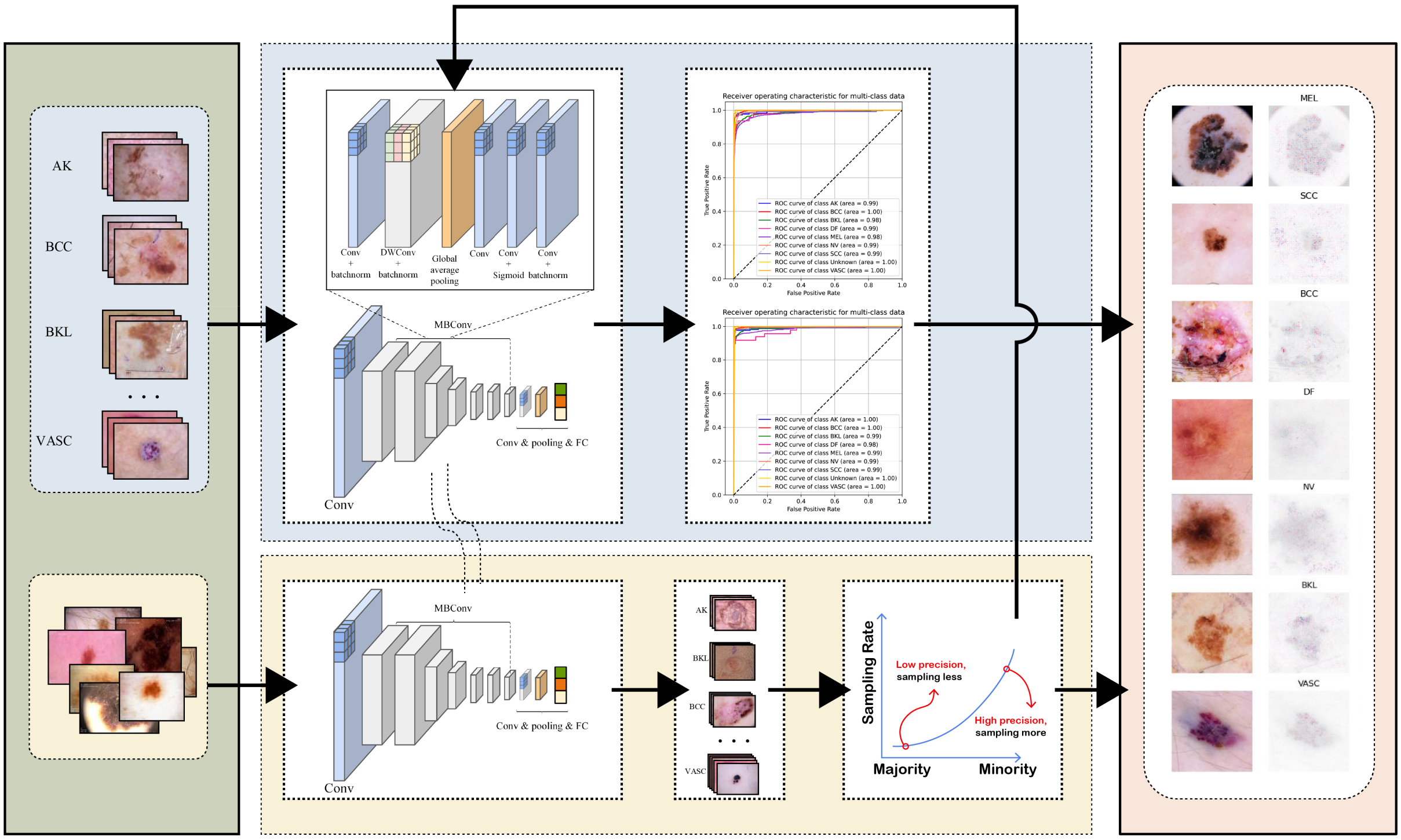

**Figure 2:** Overview of the proposed ISDL framework

## 2.3. Image augmentation

Limited supervised training is an inescapable problem in cross-deep learning, particularly in the medical field (Kostrikov et al., 2020; Shorten and Khoshgoftaar, 2019). The most common method for reducing an overfitting of the image data is to transform the original images without complex computations (Krizhevsky et al., 2012; Ha et al., 2020). Several effective image augmentation techniques have been proposed for computer vision (Chen et al., 2020). These studies successfully countered an overfitting in output-invariant tasks by applying a transformation to the input image. For skin diagnosis classification, we utilised eight types of transformation: zoom, padding, random rotation, random cut-out, random flip, transpose, random crop, and random brightness.

## 2.4. Semi-supervised learning model

This study is aimed at addressing the class imbalance problem in a semi-supervised case of an intelligent-aided dermatological diagnosis. That is, we have certain labelled samples and unmarked samples, and the distribution of these samples is a class imbalance. Inspired by Wei et al. (2021), we exploited the advantage of the high precision of the model for a minority class and adopted a self-training method. Pseudo-labelled samples from minority classes are chosen with a higher probability based on the estimated class distribution at each iteration. Furthermore, a semi-supervised learning framework for skin disease class imbalance was established. The algorithm is described as follows:

Self-training is an iterative strategy commonly used in a semi-supervised learning framework (Scudder, 1965). We denote the pseudo-labelled dataset as $\hat{Z} = \{(z_m, \hat{y}_m) : m \in (1, \cdots, M)\}$, where pseudo-labels $\hat{y}_m$ are generated through the predictions of the teacher model. In general, we suppose that the classes are sorted in descending order based on the cardinality. e.t, $N_1 \geq N_2 \geq \cdots \geq N_L$. Following the setting in Wei et al. (2021), in the first step, we apply semi-supervised learning (SSL) methods to exploit both labelled and unlabelled datasets to produce a better teacher model instead of solely training on labelled data. Moreover, a subset of pseudo-labelled datasets is selected to expand the labelled data, $X' = X \cup \hat{S}$. $\hat{S} \subset \hat{Z}$, i.e., $X' = X \cup \hat{S}$. We select $\hat{S}$ following a class-rebalancing rule, i.e., the smaller the sample size of class $l$, the more unlabelled instances predicted as class $l$ that are selected into the pseudo-labelled set $\hat{S}$. Unlabelled samples that are predicted as class $l$ are included into $\hat{S}$ at the rate of $z_l = \left(\frac{N_{L+1-l}}{N_1}\right)^{\alpha}$, where $\alpha \geq 0$ tunes the sampling rate and thus the size of $\hat{S}$. Minority class samples have a significantly higher precision than majority class samples, which means that once such a prediction is made, we can be confident that it is correct. Therefore, when selecting pseudo-labelled samples at each iteration, we choose those with the greatest confidence. In addition, we employed the progressive distribution alignment technique (Berthelot et al., 2019) with the following parameters: $\alpha = 1/10$.

## 2.5. Teacher model

An effective teacher model is of great significancesignification for the semi-supervised learning of class rebalancing. Model scaling has provenbeen proved to be an essen-tial strategy for improving the performance of convolutional neural networks (CNNs) in many studies, suchresearch, for instance as, network depth (He et al., 2016), network width (Zagoruyko and Komodakis, 2016), and image resolution (Huang et al., 2019). However, due owing to the huge significantly large combination space , a manual adjustment is challenging. In this paper, an effective CNN model, EfficientNet, is introduced to build the teacher model (Tan and Le, 2019). Then, aAn skin disease intelligent skin disease diagnosis model wasis established, with the model architecture based on mobile-size baseline convolution (MBConv). The MBConv consists of 32 layers of a 3 × 3 convolution kernel, 16 MBConv6, $k3 \times 3$; 24 MBConv6, $k3 \times 3$; 40 MBConv6, $k5 \times 5$; 80 MBConv6, $k3 \times 3$; 112 MBConv6, $k5 \times 5$; 192 MBConv6, $k5 \times 5$; 320 MBConv6, $k3 \times 3$; a 1280 1 × 1 convolution layer of the convolution kernel, and an average pooling layer, and a full connection layer with an nine9 outputs. The structures of the model and the mobile size baseline size are listed in Tablesshown in Table 2 and Table 3, respectively. ParameterThe parameter k of each mobile size baseline size refers to the channelchannels valuesvalue listed in Table 2.

**Table 2**
Structure of EfficientNet

| Stage | Operator | Resolution | Channels | Layers |
|---|---|---|---|---|
| 1 | Conv3 × 3 | 224 × 224 | 32 | 1 |
| 2 | MBConv6,$k3 \times 3$ | 112 × 112 | 16 | 1 |
| 3 | MBConv6,$k3 \times 3$ | 112 × 112 | 24 | 2 |
| 4 | MBConv6,$k5 \times 5$ | 56 × 56 | 40 | 2 |
| 5 | MBConv6,$k3 \times 3$ | 28 × 28 | 80 | 3 |
| 6 | MBConv6,$k5 \times 5$ | 14 × 14 | 112 | 3 |
| 7 | MBConv6,$k5 \times 5$ | 14 × 14 | 192 | 4 |
| 8 | MBConv6,$k3 \times 3$ | 7 × 7 | 320 | 1 |
| 9 | Conv1 × 1&Pooling &FC | 7 × 7 | 1280 | 1 |

**Table 3**
Structure of Mobile-size baseline

| Stage | Operator | Filters | Kernel size |
|---|---|---|---|
| 1 | Conv2D | in channels×1.0 | 1 × 1 |
| 2 | BatchNormalization | – | – |
| 3 | DepthwiseConv2D | 1 | $k \times k$ |
| 4 | BatchNormalization | – | – |
| 5 | GlobalAveragePooling2D<br>Conv2D<br>Conv2D<br>Sigmoid | in channels×1.0 | 1 × 1 |
| 6 | Conv2D | | 1 × 1 |
| 7 | BatchNormalization | – | – |
| 8 | Dropout, 0.2 | – | – |

## 2.6. Model interpretation

In this study, we introduce the SHAP method to explain our deep learning model (Lundberg et al., 2019). The Shapley value approach decomposes the entire model into linear combinations of all model configurations trained by each possible combination of explanatory factor features (Bussmann et al., 2021). The SHAP computational framework can be used to create a Shapley value-based method (Lundberg and Lee, 2017), which is an additive interpretative model inspired by cooperative game theory. All features are regarded as contributors and have been proven to be consistent with the importance of features in theory. The prediction results of the optimal machine learning model can be interpreted using the SHAP method; that is, we can measure the contribution of each pixel in the skin disease image to the outcome.

SHAP defines an interpretation as

$$g\left(z'\right) = \phi_0 + \sum^{M} \phi_j z'_j, \tag{1}$$

where $g$ denotes the interpretation model, $z' \in \{0,1\}^M$ is the alliance vector, and $M$ denotes the size of the largest alliance. The Shapley value of feature $j$ is recorded as $\phi_j \in \mathbb{R}$. In the alliance vectors, 1 indicates that the corresponding feature exists, whereas 0 indicates that it does not. As for the interested instance $x$, all of the alliance vectors are equal to 1, which means that features exist. It can then be simplified as follows:

$$g\left(x'\right) = \phi_0 + \sum^{M} \phi_j. \tag{2}$$

The Shapley value is calculated by SHAP, and the following SHAP kernel is established:

$$\pi_x\left(z'\right) = \frac{(M-1)}{\begin{pmatrix} M \\ |z'|' \end{pmatrix} |z'| \,(M - \mid z' \mid)} \tag{3}$$

where $|z'|$ denotes the number of current features in instance $z'$.

Then a weighted linear regression model is established:

$$g\left(z'\right) = \phi_0 + \sum_{j=1}^{M} \phi_j z_j \tag{4}$$

The linear model $g$ is trained by optimizing the following optimization function $L$

$$L\left(f, g, \pi_x\right) = \sum_{z'} \left[f\left(h_x\left(z'\right)\right) - g\left(z'\right)\right]^2 \pi_x\left(z'\right) \tag{5}$$

where $L(.)$ is an optimization function, $f(.)$ represents the predicted value of the interpreted model, $h(.)$ denotes the mapping function, that is, the illustrative function $z'$ (0 or 1) is mapped to the corresponding value of the instance $x$ to be interpreted; $Z$ is the training data. By optimizing the square sum of errors of the linear model, the estimation coefficient $\phi_j$ of the model is the Shapley value to be solved.

**Table 4**
Classification performance of different models on testing data without duplicate removal strategy.

| Model | Sensitivity | | Specificity | | Macro-F1 | | AUC | | Paras |
|---|---|---|---|---|---|---|---|---|---|
| | naive | ISDL | naive | ISDL | naive | ISDL | naive | ISDL | |
| VGGNet | 0.864±0.0053 | 0.876±0.0121 | 0.830±0.0079 | 0.852±0.0083 | 0.844±0.0049 | 0.862±0.0105 | 0.992±0.0002 | 0.994±0.0001 | 134.3 |
| Inception | 0.855±0.0067 | 0.876±0.0082 | 0.852±0.0051 | 0.870±0.0010 | 0.853±0.0044 | 0.861±0.0131 | 0.991±0.0002 | 0.994±0.0007 | 21.82 |
| DenseNet | 0.662±0.0073 | 0.678±0.0154 | 0.578±0.0060 | 0.588±0.0055 | 0.602±0.0068 | 0.616±0.0052 | 0.973±0.0002 | 0.972±0.0005 | 12.7 |
| EfficientNet | **0.940**±0.0137 | **0.975**±0.0107 | **0.940**±0.0060 | **0.973**±0.0034 | **0.939**±0.0104 | **0.974**±0.0072 | **0.998**±0.0001 | **0.999**±0.0001 | 7.2 |

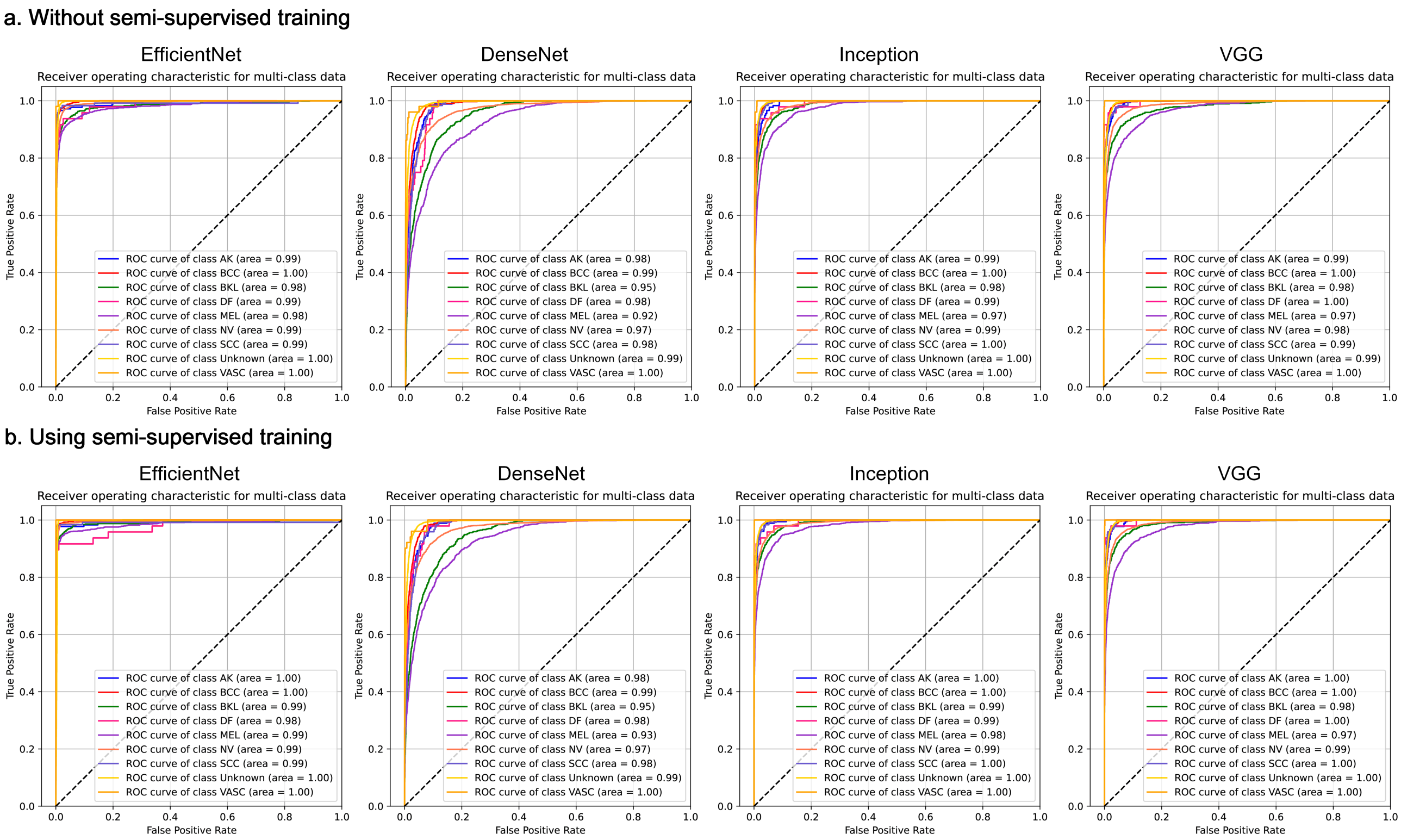

**Figure 3:** Performance of ROC curves of different models using semi-supervised training mechanism.

# 3. Criteria and details

## 3.1. Evaluation criteria

The generalisation ability of the final model was evaluated by testing its prediction performance on the testing dataset. We used different evaluation methods and conducted extensive experimental studies to compare their performance, including the accuracy, precision, sensitivity, and specificity. We denote true positive, true negative, false positive, and false negative as TP, TN, FP, and FN, respectively. The calculation method for each performance indicator is as follows:

**Accuracy:** Accuracy is the proximity of the measurement results to the true value.

$$\text{Accuracy} = \frac{TP+TN}{TP+FN+FP+TN} \tag{6}$$

**Sensitivity:** Sensitivity (recall) indicates the proportion of a test correctly generating a positive outcome.

$$\text{Sensitive} = \frac{TP}{TP+FN} \tag{7}$$

**Specificity:** Specificity measures the ability of a test to correctly identify people without a disease.

$$\text{Specificity} = \frac{TN}{FP+TN} \tag{8}$$

**Macro-F1 score:** The Macro F1 score metric is frequently used to measure multi-class classification problems.

$$\text{Macro} - F1 = \frac{1}{L}\sum_{i=1}^{L}\frac{\text{recall}_i \times \text{precison}_i}{\text{recall}_i + \text{precison}_i} \times 2, \tag{9}$$

where $L$ denotes the number of categories.

In addition, the area under the subject working characteristic curve (AUC) is widely used as an evaluation index of the prediction model (Fawcett, 2004). This method combines the true positive and false negative rates to measure the prediction ability of the model and considers the discrimination ability of the model for positive and negative cases. An AUC of 1 indicates that the model is a perfect prediction model, whereas an AUC of 0.5 means that the model is a random prediction model.

### 3.2. Implementation detail

During the training phase, the dataset was randomly divided into three parts: training, validation, and testing sets, with ratios of 80%, 10%, and 10%, respectively. Moreover, we maintained the corresponding proportions of different skin illnesses during the random partitioning phase to ensure that the framework did not "bias" the learning after the split.

For the classification network, an adaptive momentum estimation (Adam) was used as the classifier optimiser (Kingma and Ba, 2014). This algorithm combines the superiority of Adagrad with the momentum gradient descent algorithm, which not only adapts to sparse gradients, it also alleviates the problem of a gradient oscillation. The initial learning rate $1e{-}4$ decayed by 0.96 every 1 epoch after 50 epochs. We also used SiLU (Swish-1) activation (Hendrycks and Gimpel, 2016; Ramachandran et al., 2017; Elfwing et al., 2018) for the convolution operation. To obtain the final predicted probability, the Softmax function was utilised in the last layer of the network. All of the above models are conducted on a Dell server with 128 GB of RAM and an RTX2080Ti GPU and implemented in Python.

## 4. Experiment analysis

### 4.1. Validation of proposed ISDL framework

VGGNet (Simonyan and Zisserman, 2014), Inception (Szegedy et al., 2016), DenseNet (Huang et al., 2017), and EfficientNet (Tan and Le, 2019) were considered benchmark models for comparison with the proposed ISDL framework. These models are classical and perform excellently in terms of image classification. Table 4 illustrates the performance of the testing dataset of the loss functions, such as the accuracy, sensitivity, specificity, macro-F1 score, and area under the receiver operating characteristic curve (AUC). Naïve means that the models did not use a semi-supervised learning mechanism. It is evident that the proposed ISDL is superior to the four benchmark models without semi-supervised learning. In terms of the sensitivity, specificity, macro-F1 scores and AUC, our ISDL produced the best performance based on EfficientNet, yielding values of 0.975, 0.973, 0.974 and 0.999, respectively. As for EfficientNet bone block, the Sensitivity, specificity and macro-F1 scores were greater than 0.964, 0.970 and 0.967 respectively. Our ISDL obtains an AUC score of above 0.999 for all testing samples. Furthermore, the parameters of VGGNet and Inception models are the largest, with parameters of 134.3 and 21.82 million. The results demonstrate that our ISDL outperforms other models based on the predictive accuracy and model efficiency.

The performance of the ROC curves of the different models using a semi-supervised training mechanism is illustrated in Figure 3. Level a denotes EfficientNet, DenseNet, Inception, and VGGNet, without using the semi-supervised learning mechanism, for the ROC curves of different skin diseases. Each type of disease is marked by a different colour, and the area in the legend denotes the AUC value. Level b shows the ROC curve and AUC value of these four benchmark prediction models using our proposed semi-supervised learning mechanism, ISDL.

The ROC curve reflects the correlation between the sensitivity and specificity of the final deep learning model, as well as the recognition ability. By controlling the threshold of the deep-learning classifier, the correlation between them can be presented in the form of a continuous curve. The closer the ROC curve is to the upper-left corner, the higher the accuracy of the curve corresponding to the disease diagnosis and the recognition ability of the model. The AUC value was defined as the area below the ROC curve. Compared to the ROC curve, the AUC value can more intuitively quantify the diagnostic ability of the model. The larger the AUC value of the model, the stronger its recognition ability. This study compares the model representation ability of EfficientNet, Inception, DenseNet, and VGGNet in cases with and without semi-supervision . Among the four models, EfficientNet performed better than the other models on the testing dataset. Compared to the other three models, the AUC value of each category was close to 1. By contrast, in the ROC curve of DenseNet, the AUC value of melanoma was approximately 0.97, which was also unsatisfactory compared with that of the other models. The recognition ability of each individual deep learning model was improved through semi-supervised learning.

To further verify the superiority of the model proposed in this paper, we compared the prediction performance of two previous methods on ISIC 2018 and ISIC 2019 data sets respectively (see Table 5). Data set category preprocessing and sample division follow the setting of the papers (Gessert et al., 2018; Kassem et al., 2020). The results show that under the same data set division, the prediction performance of our ISDL and ISDLplus are better than that of the existing methods.

T-distributed stochastic neighbour embedding (t-SNE) is a nonlinear dimensionality-reduction algorithm. It excels at revealing the local structure in high-dimensional data and is extremely suitable for the visualisation of a high-dimensional data reduction into two or three dimensions (Van der Maaten and Hinton, 2008; Kobak and Berens, 2019; Linderman et al., 2019). Using t-SNE, we visualised the learned features of the deep learning model on a test subset of skin diseases (see Figure 6). Each type of marker represents a corresponding type of disease, the captured features for nine skin diseases in the samples are well separated.

### 4.2. Interpretability of our ISDL

Figure 4 provides an illustrative interpretation of dermatological images using the SHAP method. Specifically, we visualize the prediction results for the top three skin disease classes with the highest probability, from left to right, given an input image of the skin disease. The results of the first instance show that the input image is of melanoma. With melanoma, the invasion of malignant melanocytes can develop upward in the epidermis alone or nest (through a Paget-like diffusion) or grow radially at the epidermal-dermal junction (DEJ). The model predicts that the image is most likely to be melanoma, followed by melanocytic naevus, and the third predicted value indicates benign keratinoid lesions, showing that our ISDL accurately diagnoses the sample as melanoma. The coloured pixels in the graph denote the contribution to the prediction made by the model, that is, the Shapley value. The red mark indicates that the pixel increases the probability of the prediction and makes a positive contribution to the prediction of the result, whereas the blue mark indicates that the pixel contributes to the prediction in a negative direction, reducing the probability. The depth of the colour indicates the extent of the contribution. The darker the colour is, the greater the contribution of the pixel at the corresponding position.

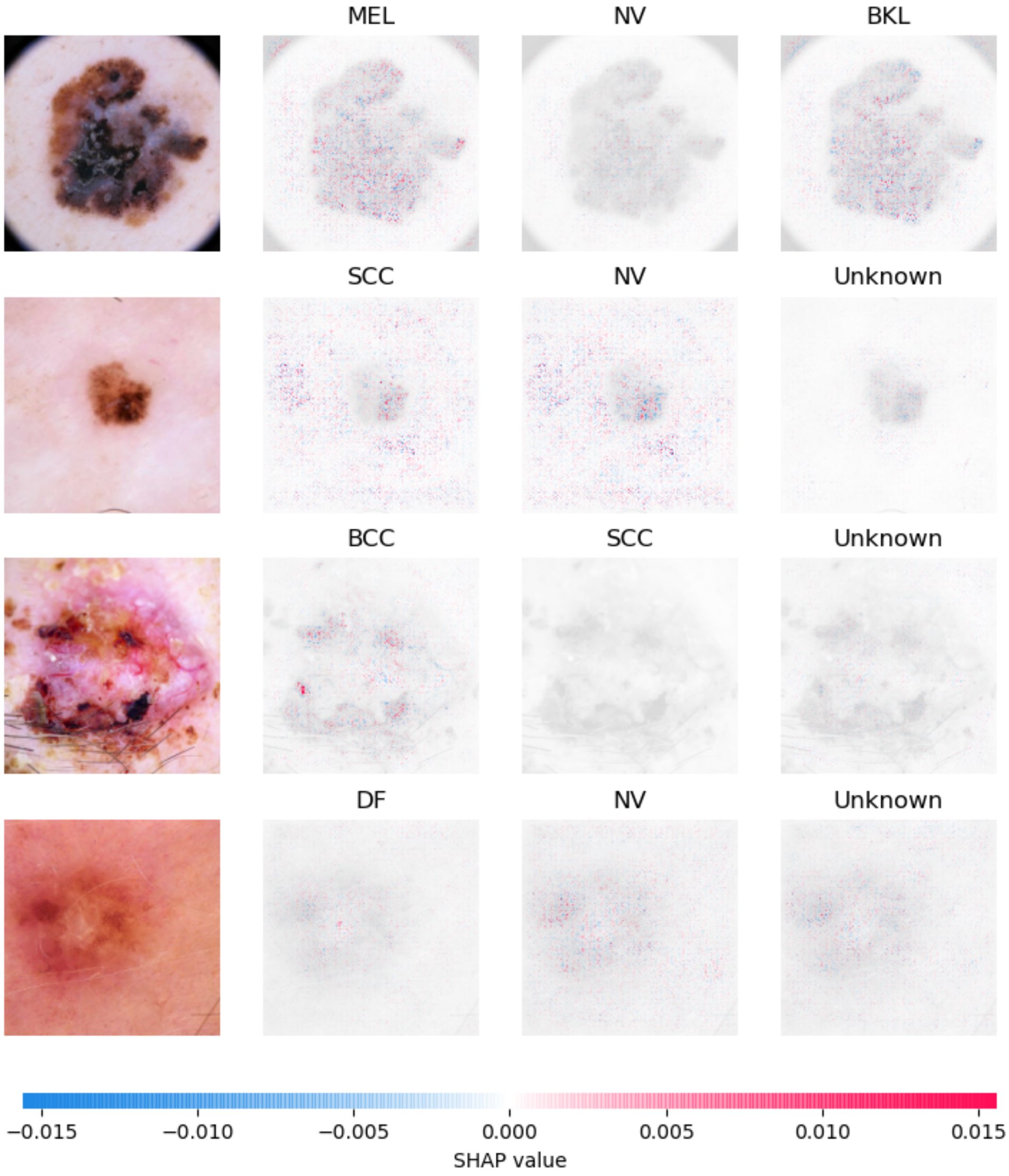

**Figure 4:** Illustrations of interpretation in dermatology image using SHAP method.

The input of the second example is a photograph of squamous-cell carcinoma (SCC), which is a precancerous and malignant keratinising tumour. Most of the images are non-pigmented, which means that their dermatoscopic diagnosis is mainly based on a vascular pattern analysis. The model accurately predicts that the maximum probability is that the photograph shows squamous-cell carcinoma, whereas the probabilities of it showing naevus or an unknown object come in second and third, respectively. The SHAP plot shows that the contribution of small areas in the middle lesion area to the prediction of squamous cell car-cinoma was negative, whereas the contribution to the prediction of nevus was positive, and the contribution of other reddish lesions around the skin to the final result was positive. This demonstrates that our ISDL diagnoses the sample as benign keratosis by capturing the features from the surrounding reddish skin rather than melanocytic naevus.

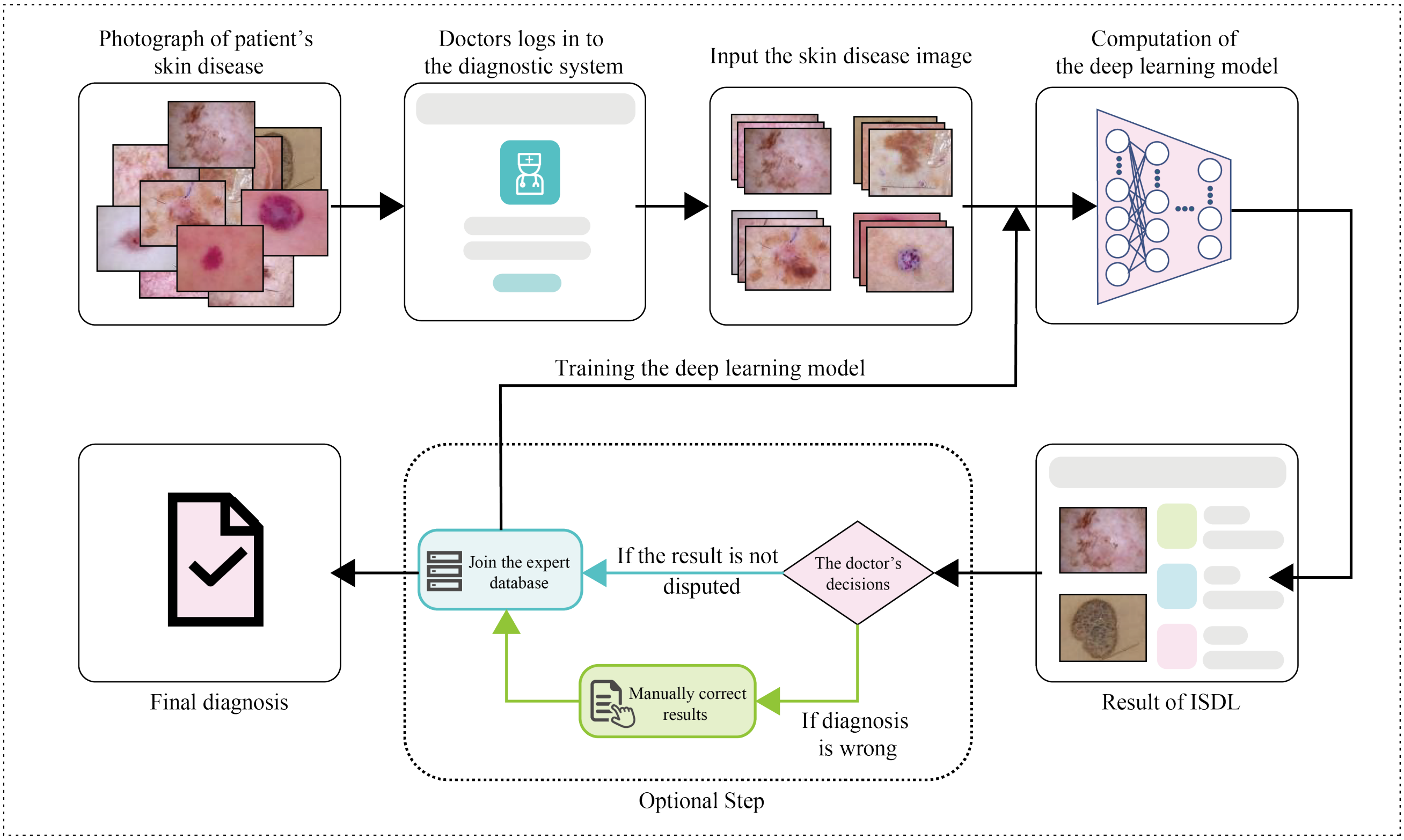

**Figure 5:** Construction of ISDL application system.

**Table 5**
Compared with previous methods on ISIC 2018 and ISIC 2019 datasets.

| | Evaluation on ISIC 2018 | | | | |
|---|---|---|---|---|---|
| Methods | Accuracy | Precison | Recall | F1 score | AUC |
| Gessert et al. | 0.851 | × | × | × | 0.976 |
| ISDL | 0.869±0.005 | 0.833±0.011 | 0.725±0.014 | 0.762±0.021 | 0.969±0.016 |
| ISDLplus | **0.901**±0.004 | **0.899**±0.011 | **0.776**±0.021 | **0.828**±0.017 | **0.980**±0.005 |
| | Evaluation on ISIC 2019 | | | | |
| Methods | Accuracy | Precison | Recall | F1 score | AUC |
| Kasem et al. | 0.9492 | 0.8036 | 0.7980 | 0.8007 | 0.9760 |
| ISDL | 0.9668±0.002 | 0.9705±0.013 | 0.9502±0.009 | 0.9598±0.009 | 0.9950±0.001 |
| ISDLplus | **0.9718**±0.001 | **0.9741**±0.002 | **0.9548**±0.005 | **0.9641**±0.003 | **0.9960**±0.001 |

The input of the third sample is an image of basal-cell carcinoma. Basal cell carcinoma is the most common type of skin cancer. Clinically, this skin cancer can present in a variety of forms, from erythema to ulcerative nodules. In terms of the prediction results, the top three were basal-cell carcinoma, squamous-cell carcinoma, and unknown. The results show that the model provides an accurate prediction. The SHAP diagram also illustrates that the model grasps the lesion area in the image, particularly the two black lesion positions at the bottom of the image, and the Shapley value of the corresponding position is significantly red. These results indicate that the model captures the characteristics of the location of black lesions and predicts the sample as basal-cell carcinoma.

The input of the fourth sample is dermatofibroma. A cutaneous fibroma is a fibrous skin lesion characterised by an increase in dermal fibroblasts. These are often mixed with varying degrees of inflammatory cells, thickened collagen bundles, and blood vessels. The central mean area may be mixed with a white pattern, a peripheral fine pigment network of fibres, and a vascular pattern . Similarly, the model accurately predicts with maximum probability that the pho-tograph shows dermatofibroma, whereas the probabilities of it showing a naevus or an unknown object rank second and third, respectively. The results indicate that on the SHAP plot predicted as dermatofibroma, the small blocks in the middle region have a significant positive contribution to the prediction, whereas the surrounding reddened regions contribute significantly to the prediction of a naevus, and the contribution of the middle region is greater. This shows that the model determines that the contribution of the middle region to the prediction of dermatofibroma is a decisive factor in comparison to the pixels corresponding to the surrounding reddish region.

**Table 6**
Classification performance of different models with duplicate removal strategy.

| Model | Sensitivity | | Specificity | | Macro-F1 | | AUC | | Paras |
|---|---|---|---|---|---|---|---|---|---|
| | naive | ISDL | naive | ISDL | naive | ISDL | naive | ISDL | |
| VGGNet | 0.760±0.0049 | 0.823±0.0004 | 0.741±0.0166 | 0.728±0.0002 | 0.748±0.0055 | 0.762±0.0010 | 0.986±0.0003 | 0.992±0.0004 | 134.3 |
| Inception | 0.651±0.0008 | 0.753±0.0026 | 0.628±0.0089 | 0.693±0.0112 | 0.633±0.0062 | 0.717±0.0074 | 0.976±0.0001 | 0.984±0.0004 | 21.82 |
| DenseNet | 0.666±0.0042 | 0.679±0.0081 | 0.665±0.0053 | 0.714±0.0128 | 0.660±0.0002 | 0.684±0.0006 | 0.976±0.0004 | 0.982±0.0006 | 12.7 |
| EfficientNet | **0.890**±0.0088 | **0.912**±0.0004 | **0.813**±0.0026 | **0.829**±0.0006 | **0.853**±0.0053 | **0.857**±0.0009 | **0.992**±0.0001 | **0.996**±0.0002 | 7.2 |

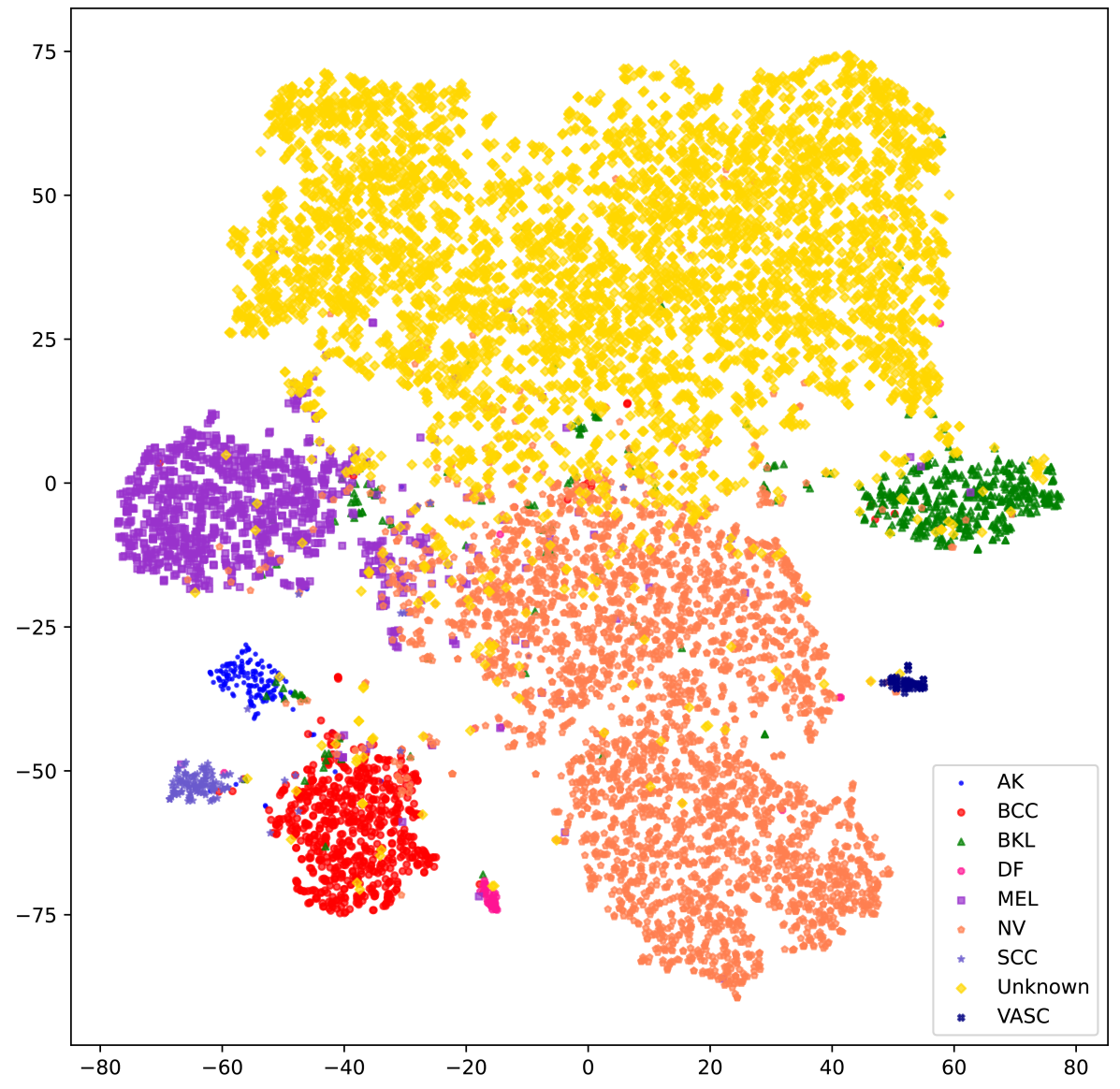

**Figure 6:** t-SNE dimensionality reduction.

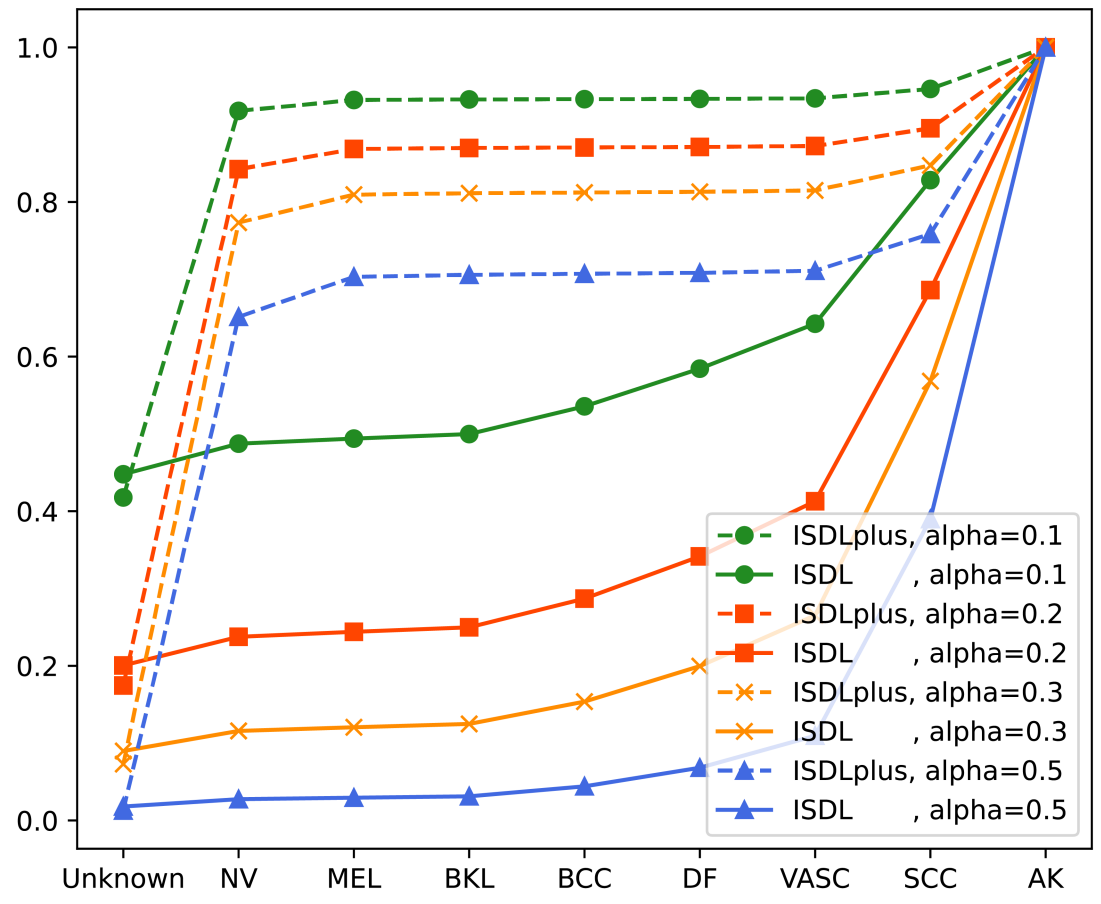

**Figure 7:** Comparison of sampling proportion between ISDL and ISDLplus.

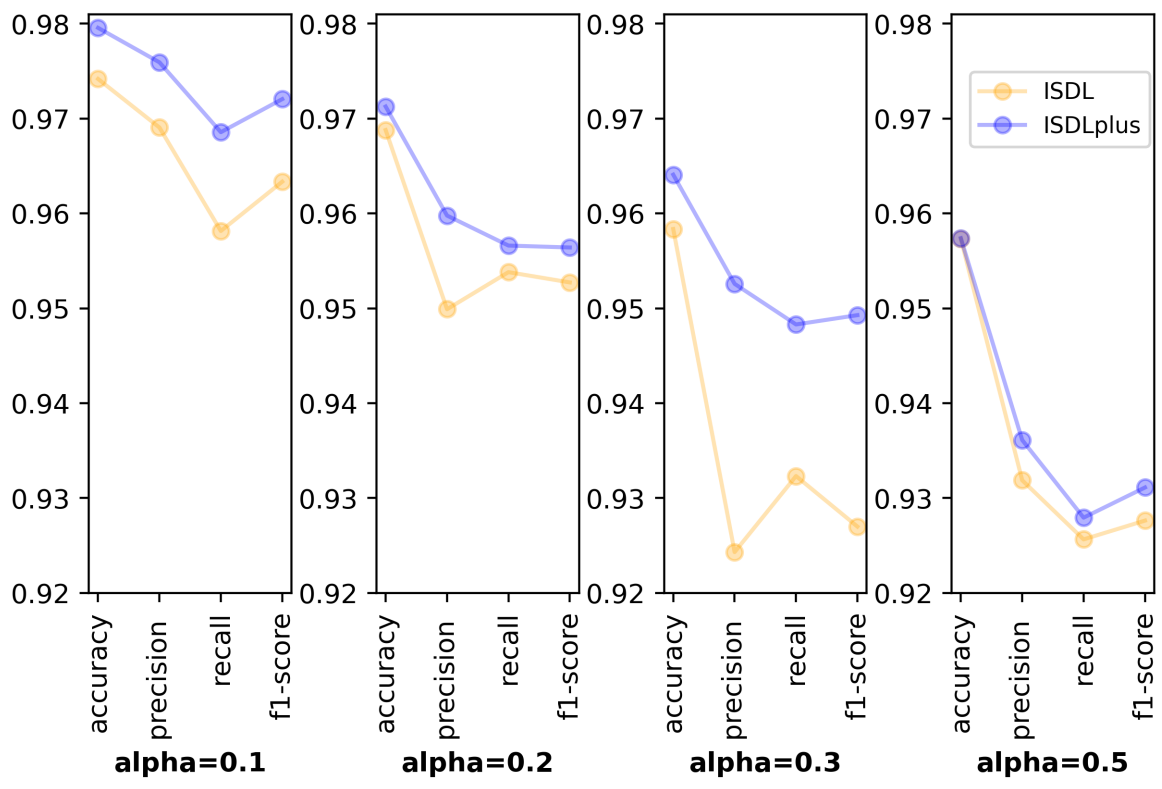

**Figure 8:** Performance comparison of ISDL and ISDLplus.

### 4.3. Skin disease intelligent diagnosis system

Based on the proposed ISDL, this study builds an intelligent skin disease diagnosis system (see Figure 5). First, it is necessary to take photographs of the patient's skin lesion area to obtain a skin disease image. The doctor then logs into the intelligent dermatological diagnosis system, uploads the dermatosis image, and uses the image as the input of the ISDL. Our ISDL predicts the probability of diagnosing each disease according to the input image and takes the disease category with the highest probability as the final diagnosis. After the disease prediction results are obtained, it is necessary to calculate the Shapley value interpretation map of the diagnosis image in combination with the skin disease image in the expert database. The reliability of a disease diagnosis can be judged in combination with the SHAP interpretation map and original input image used to obtain the final diagnosis results. In addition, the diagnosis results can also be selectively submitted to an expert as a background of the system, who will judge the diagnostic results again. If the diagnosis of the deep learning model is incorrect, the expert will correct the results, upload the final results onto the expert database, and provide the expert diagnosis results as feedback. If the model diagnosis is correct, it will be directly added to the expert database and again provide the expert diagnosis results as feedback. When the expert database established through a disease diagnosis accumulates a certain number of samples, the expert database and the training database of the original deep learning model are used for model retraining to continuously improve the accuracy and stability of the deep learning model diagnosis. The above results demonstrate that our ISDL can effectively interpret the prediction results obtained by the model and quantify the contribution at the pixel level in terms of direction and size.

**Table 7**
Accuracy performance of top 1 and top 2 conducted on different duplicate removal strategy

| | Model | without duplicate removal strategy | | with duplicate removal strategy | |
|---|---|---|---|---|---|
| | | naive | ISDL | naive | ISDL |
| Top 1 (%) | VGGNet | 91.79±0.10 | 92.50±0.47 | 89.72±0.25 | 91.80±0.03 |
| | Inception | 92.20±0.10 | 93.40±0.50 | 85.81±0.12 | 88.57±0.16 |
| | DenseNet | 84.68±0.20 | 85.11±0.14 | 86.37±0.49 | 87.71±0.19 |
| | EfficientNet | **95.84**±0.41 | **97.90**±0.23 | **93.14**±0.02 | **94.07**±0.01 |
| Top 2 (%) | VGGNet | 98.08±0.09 | 98.13±0.14 | 97.05±0.06 | 98.00±0.03 |
| | Inception | **98.21**±0.02 | 98.87±0.01 | 95.02±0.08 | 96.92±0.03 |
| | DenseNet | 94.31±0.18 | 94.46±0.09 | 95.44±0.13 | 96.42±0.01 |
| | EfficientNet | 97.95±0.25 | **99.52**±0.04 | **98.53**±0.08 | **98.60**±0.05 |

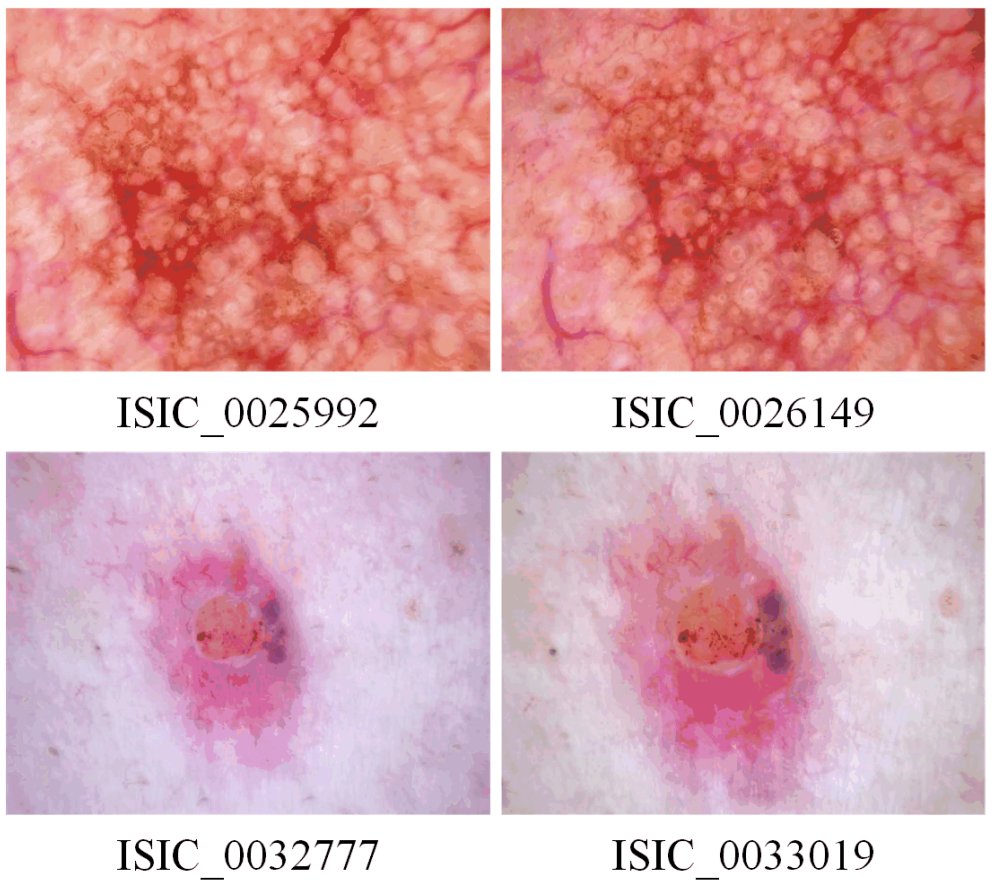

**Figure 9:** Examples of duplicate image with different filenames.

## 4.4. Sampling distribution optimization

In the previous ISDL framework, unlabelled samples predicted as class $l$ are selected by $z_l = \left(\frac{N_{L+1-l}}{N_1}\right)^{\alpha}$. The sampling proportion of each class of samples is only related to the position of the pseudo-labelled samples in descending order and is not related to the current number of pseudo-labelled samples of this class. Therefore, when there are small numbers of pseudo-labels, and such numbers are similar, the sampling proportion of these categories will differ, which is not an optimal choice. On this basis, we optimized the sampling proportion formula as follows:

$$z_l = \left(\frac{N_{L+1-l} + (N_1 - N_l)}{2N_1}\right)^{\alpha}, \quad (10)$$

where $N_1$ is the number of samples corresponding to the majority class in the pseudo-labelled dataset, and $N_l$ denotes the number of samples corresponding to the minority class in the same dataset. We distinguish this from ISDLplus. This formula not only considers the relationship between the number and location of pseudo-labelled samples, it also considers the number of current categories. Compared with ISDL, ISDLplus has fewer samples in the majority classes and can balance the number of samples in a "complementary" form to a greater extent. Under the premise of reducing the rate of imbalance, a minority number of pseudo-labelled classes that can be fully included as much as possible is promoted.

Figure 7 illustrates the change in the control alpha parameter on the sampling proportion of pseudo-labelled skin disease samples under the two proposed frameworks, ISDL and ISDLplus. In general, the alpha parameter controls the total number of samples from the pseudo-labelled dataset. The lower the alpha value, the greater the number of population samples, which means that more pseudo-labelled samples are included in the training set. For the same alpha value, the proportion of samples of ISDLplus being in the majority of unknown classes is less than that of ISDL, whereas the proportion of samples of other minority classes is higher than that of ISDL and thus more minority classes are fully utilised . Therefore, the overall imbalance rate of ISDLplus is lower than that of ISDL. Figure 8 shows the performances of ISDL and ISDLplus on the dermatological testing dataset under different alpha values. The performance of ISDLplus is superior to that of ISDL in terms of accuracy, F1 score, precision, and recall.

**Table 8**
Summary of duplicate image files.

| Year | Train set | Test set |
|---|---|---|
| 2018 | 10,015 | 1,512 |
| 2019 | 1,094 | 2 |
| 2020 | 433 | 78 |
| Total | 11,542 | 78 |

### 4.5. Further experiment with duplicate removal strategy

ISIC datasets contain tens of thousands of skin images together with gold-standard lesion diagnosis metadata which has resulted in majorcontributions to the field. However, after examining the data, we found that there were duplicate images in the data set. Sevaral illustrations of duplicate image with different filenames are shown in Figure 9. ISIC_0025992 and ISIC_0032777 found in the 2018 trianing set, and the corresponding ISIC_0026149 and ISIC_0033019 are found in the 2019 training set. Inspire the Cassidy et al. (2022), we perform the following strategy on the ISIC 2018-2020 dataset:

- Remove all duplicate binary indentical image files across all training sets (2018-2020).
- Remove al image files from each individual training set where a duplicate is found in any of the testing set.

Table 8 demonstrates the summary of duplicate skin image files. A total of 11,542 images are removed from train sets and 78 images from test sets.

Furthermore, we perform a series of benchmark tests on the data set. Training, validation and testing sets are also randomly split by 8:1:1. Stratified sampling is conducted to ensure that the imbalance rate of the dataset is consistent. As shown in Table 6, our ISDL is obviously superior to the other banchmark models in term of sensitivity, specificity, macro-F1 and AUC scores, yielding the avarage values of 0.912, 0.829, 0.857 and 0.996, respectively. As for these score performances, VGGNet performs better than other two models. However, the parameters of VGGNet is the largest, which means that a great of computational consumption. To sum up, our ISDL achieves the promissing performance in terms of the predicition and efficiency performance.

In addition, we compare the top1 and top2 accuracy scores conducted on different duplicate removal strategies. On the whole, the performance of the models is predictably inferior to those without a duplicate removal strategy. However, our ISDL is still outperformed the other benchmark models. In terms of top1 accuracy scores of these two datasets (without duplicate removal strategy and with duplicate removal strategy), our ISDL obtains the average values of 97.90 and 94.07, respectively. In terms of the top2 accuracy performance, our ISDL yeilds a promising result of more than 98% accuracy.

## 5. Discussion

To the best of our knowledge, this is the first report demonstrating a semi-supervised learning framework for studying interpretability and class imbalance problems in the intelligent diagnosis of skin diseases. Machine learning has yielded remarkable achievements in many fields, such as time series regression, image recognition, speech recognition, etc (Weng et al., 2021; Schiff et al., 2022; Weng et al., 2022; Le Goallec et al., 2022; Bileschi et al., 2022; Baek and Baker, 2022). Especially, deep learning models have achieved a promising performance that is not inferior to that of dermatologists in the field of intelligent skin disease diagnosis. However, few studies have focused on the following challenges: 1) the limitation of the labelled dataset, 2) the class imbalance problem, 3) the efficiency of the deep learning model, and 4) the interpretability of deep learning. This study attempts to solve these problems by constructing an effective semi-supervised learning framework.

This study validates our proposed ISDL framework using a publicly available dermatology dataset, ISIC. By integrating the 2018-2020 dataset, a total of 58,457 skin disease images were included in the study, of which 10,857 were unlabelled images. The EfficientNet network was utilised as the teaching model in the training phase, and was trained using labelled skin images. Subsequently, in each iteration of self-training, the teacher model was used to predict the unlabelled data to obtain the pseudo-label. The samples predicted to be a minor class are then selected with a higher probability and combined into the labelled sample set. Thus, an unlabelled dataset can be effectively used to address the class imbalance problem. Our ISDL yielded an excellent performance, with an accuracy of 0.979, sensitivity of 0.975, specificity of 0.973, macro-F1 score of 0.974 and AUC of 0.999 for multi-label skin disease classification based on the EfficientNet model. Moreover, considering the relative position of the number of samples, we propose a sampling distribution optimisation strategy for selecting pseudo-labelled samples in a more effective manner. Compared to ISDL, ISDLplus can select pseudo-labelled datasets more effectively and reduce the imbalance rate. The results demonstrate that ISDLplus performs better than ISDL in terms of different evaluation criteria. Furthermore, we conduct a duplicate removal strategy to generate a new skin image dataset. The prediction performance on the test set shows that our ISDL is robust.

In addition, deep learning on how to make predictions has always been a black-box problem that hinders its application in clinical practice. Combined with the SHAP method, this study proposed a model-independent interpretable method for deep learning. This method can explain the prediction of a model at the pixel level. Therefore, clinicians can know from the ISDL which pixel regions the model uses to make the disease diagnosis. The experiment results showed that the ROI quantified by the ISDL is consistent with the knowledge of dermatologists.

Our study has several limitations. First, in terms of skin disease image augmentation, we primarily used geometric and rotational methods. In future research, we will consider more image augmentation methods, particularly data enhancement methods for medical images. Second, a single modal skin image was considered in our ISDL. Further studies should include patient information and other modal data to optimise the predicted model. Finally, for the design of the system, more functions, such as quantifying the size of the lesion area, can be added.

In conclusion, our ISDL can effectively use unlabelled skin disease datasets to improve the prediction performance of the model, which will significantly reduce the labelling tasks of medical workers. Furthermore, the prediction of minority samples achieves a high precision, which means that once a sample is predicted as a minority by the teacher model, there is a higher probability of it being of a minor class. Therefore, during each iteration of self-training, the samples with high confidence are minor classes, which are selected into the labelled dataset with a higher probability for solving the class imbalance problem. In terms of deep learning interpretability, as a theoretical feature attribution method, the SHAP approach can effectively quantify the contribution of features to the prediction results at the pixel level. Our ISDL facilitates the implementation of telemedicine, particularly in remote areas. Simultaneously, ISDLs can be applied to other medical-assisted diagnostic cases.

## Acknowledgments

The authors gratefully acknowledge the financial support provided by the Major Project of the National Social Science Foundation (20&ZD137) and the National Natural Science Foundation of China (12026239).

## References

Baek, M., Baker, D., 2022. Deep learning and protein structure modeling. Nature methods 19, 13–14.

Bengio, Y., Courville, A., Vincent, P., 2013. Representation learning: A review and new perspectives. IEEE transactions on pattern analysis and machine intelligence 35, 1798–1828.

Berthelot, D., Carlini, N., Cubuk, E.D., Kurakin, A., Sohn, K., Zhang, H., Raffel, C., 2019. Remixmatch: Semi-supervised learning with distribution alignment and augmentation anchoring. arXiv preprint arXiv:1911.09785 .

Bileschi, M.L., Belanger, D., Bryant, D.H., Sanderson, T., Carter, B., Sculley, D., Bateman, A., DePristo, M.A., Colwell, L.J., 2022. Using deep learning to annotate the protein universe. Nature Biotechnology , 1–6.

Bussmann, N., Giudici, P., Marinelli, D., Papenbrock, J., 2021. Explainable machine learning in credit risk management. Computational Economics 57, 203–216.

Cassidy, B., Kendrick, C., Brodzicki, A., Jaworek-Korjakowska, J., Yap, M.H., 2022. Analysis of the isic image datasets: usage, benchmarks and recommendations. Medical image analysis 75, 102305.

Chan, T.H., Jia, K., Gao, S., Lu, J., Zeng, Z., Ma, Y., 2015. Pcanet: A simple deep learning baseline for image classification? IEEE transactions on image processing 24, 5017–5032.

Chen, T., Kornblith, S., Norouzi, M., Hinton, G., 2020. A simple framework for contrastive learning of visual representations, in: International conference on machine learning, PMLR. pp. 1597–1607.

Codella, N., Rotemberg, V., Tschandl, P., Celebi, M.E., Dusza, S., Gutman, D., Helba, B., Kalloo, A., Liopyris, K., Marchetti, M., et al., 2019. Skin lesion analysis toward melanoma detection 2018: A challenge hosted by the international skin imaging collaboration (isic). arXiv preprint arXiv:1902.03368 .

Commission, N.H., et al., 2020. National health commission of the people's republic of china. The Guideline on Diagnosis and Treatment of the Novel Coronavirus Pneumonia (NCP): Revised Version of the 7th Edition .

Elfwing, S., Uchibe, E., Doya, K., 2018. Sigmoid-weighted linear units for neural network function approximation in reinforcement learning. Neural Networks 107, 3–11.

Fawcett, T., 2004. Roc graphs: Notes and practical considerations for researchers. Machine learning 31, 1–38.

Garcia-Garcia, A., Orts-Escolano, S., Oprea, S., Villena-Martinez, V., Garcia-Rodriguez, J., 2017. A review on deep learning techniques applied to semantic segmentation. arXiv preprint arXiv:1704.06857 .

Gessert, N., Sentker, T., Madesta, F., Schmitz, R., Kniep, H., Baltruschat, I., Werner, R., Schlaefer, A., 2018. Skin lesion diagnosis using ensembles, unscaled multi-crop evaluation and loss weighting. arXiv preprint arXiv:1808.01694 .

Goceri, E., 2019. Skin disease diagnosis from photographs using deep learning, in: ECCOMAS thematic conference on computational vision and medical image processing, Springer. pp. 239–246.

Ha, Q., Liu, B., Liu, F., 2020. Identifying melanoma images using efficientnet ensemble: Winning solution to the siim-isic melanoma classification challenge. arXiv preprint arXiv:2010.05351 .

Habif, T.P., Chapman, M.S., Dinulos, J.G., Zug, K.A., 2017. Skin Disease E-Book: Diagnosis and Treatment. Elsevier Health Sciences.

Hameed, A., Umer, M., Hafeez, U., Mustafa, H., Sohaib, A., Siddique, M.A., Madni, H.A., 2021. Skin lesion classification in dermoscopic images using stacked convolutional neural network. Journal of Ambient Intelligence and Humanized Computing , 1–15.

He, K., Zhang, X., Ren, S., Sun, J., 2016. Deep residual learning for image recognition, in: Proceedings of the IEEE conference on computer vision and pattern recognition, pp. 770–778.

Hendrycks, D., Gimpel, K., 2016. Gaussian error linear units (gelus). arXiv preprint arXiv:1606.08415 .

Huang, G., Liu, Z., Van Der Maaten, L., Weinberger, K.Q., 2017. Densely connected convolutional networks, in: Proceedings of the IEEE conference on computer vision and pattern recognition, pp. 4700–4708.

Huang, Y., Cheng, Y., Bapna, A., Firat, O., Chen, D., Chen, M., Lee, H., Ngiam, J., Le, Q.V., Wu, Y., et al., 2019. Gpipe: Efficient training of giant neural networks using pipeline parallelism. Advances in neural information processing systems 32, 103–112.

Iqbal, J., 2021. Dermatologist-level classification of skin cancer with deep neural networks .

Janoria, H., Minj, J., Patre, P., 2021. Classification of skin disease using traditional machine learning and deep learning approach: A review, in: Intelligent Data Communication Technologies and Internet of Things: Proceedings of ICICI 2020, Springer. pp. 643–651.

Kassem, M.A., Hosny, K.M., Fouad, M.M., 2020. Skin lesions classification into eight classes for isic 2019 using deep convolutional neural network and transfer learning. IEEE Access 8, 114822–114832.

Kingma, D.P., Ba, J., 2014. Adam: A method for stochastic optimization. arXiv preprint arXiv:1412.6980 .

Kobak, D., Berens, P., 2019. The art of using t-sne for single-cell transcriptomics. Nature communications 10, 1–14.

Korot, E., Guan, Z., Ferraz, D., Wagner, S.K., Zhang, G., Liu, X., Faes, L., Pontikos, N., Finlayson, S.G., Khalid, H., et al., 2021. Code-free deep learning for multi-modality medical image classification. Nature Machine Intelligence 3, 288–298.

Kostrikov, I., Yarats, D., Fergus, R., 2020. Image augmentation is all you need: Regularizing deep reinforcement learning from pixels. arXiv preprint arXiv:2004.13649 .

Krizhevsky, A., Sutskever, I., Hinton, G.E., 2012. Imagenet classification with deep convolutional neural networks. Advances in neural information processing systems 25, 1097–1105.

Le Goallec, A., Diai, S., Collin, S., Prost, J.B., Vincent, T., Patel, C.J., 2022. Using deep learning to predict abdominal age from liver and pancreas magnetic resonance images. Nature Communications 13, 1–13.

LECUN, 2015. Y, bengio y, hinton g. deep learning. Nature 521, 436–444.

Li, H., Pan, Y., Zhao, J., Zhang, L., 2020. Skin disease diagnosis with deep learning: a review. arXiv preprint arXiv:2011.05627 .

Linderman, G.C., Rachh, M., Hoskins, J.G., Steinerberger, S., Kluger, Y., 2019. Fast interpolation-based t-sne for improved visualization of single-cell rna-seq data. Nature methods 16, 243–245.

Liu, Y., Jain, A., Eng, C., Way, D.H., Lee, K., Bui, P., Kanada, K., de Oliveira Marinho, G., Gallegos, J., Gabriele, S., et al., 2020. A deep learning system for differential diagnosis of skin diseases. Nature medicine 26, 900–908.

Lundberg, S.M., Erion, G., Chen, H., DeGrave, A., Prutkin, J.M., Nair, B., Katz, R., Himmelfarb, J., Bansal, N., Lee, S.I., 2019. Explainable ai for trees: From local explanations to global understanding. arXiv preprint arXiv:1905.04610 .

Lundberg, S.M., Lee, S.I., 2017. A unified approach to interpreting model predictions, in: Proceedings of the 31st international conference on neural information processing systems, pp. 4768–4777.

Van der Maaten, L., Hinton, G., 2008. Visualizing data using t-sne. Journal of machine learning research 9.

Madooei, A., Drew, M.S., 2016. Incorporating colour information for computer-aided diagnosis of melanoma from dermoscopy images: A retrospective survey and critical analysis. International journal of biomedical imaging 2016.

Ouahabi, A., Taleb-Ahmed, A., 2021. Deep learning for real-time semantic segmentation: Application in ultrasound imaging. Pattern Recognition Letters 144, 27–34.

Ramachandran, P., Zoph, B., Le, Q.V., 2017. Searching for activation functions. arXiv preprint arXiv:1710.05941 .

Saez, A., Acha, B., Serrano, C., 2014. Pattern analysis in dermoscopic images, in: Computer vision techniques for the diagnosis of skin Cancer. Springer, pp. 23–48.

Schiff, L., Migliori, B., Chen, Y., Carter, D., Bonilla, C., Hall, J., Fan, M., Tam, E., Ahadi, S., Fischbacher, B., et al., 2022. Integrating deep learning and unbiased automated high-content screening to identify complex disease signatures in human fibroblasts. Nature Communications 13, 1–13.

Scudder, H., 1965. Probability of error of some adaptive pattern-recognition machines. IEEE Transactions on Information Theory 11, 363–371.

Shorten, C., Khoshgoftaar, T.M., 2019. A survey on image data augmentation for deep learning. Journal of Big Data 6, 1–48.

Simonyan, K., Zisserman, A., 2014. Very deep convolutional networks for large-scale image recognition. arXiv preprint arXiv:1409.1556 .

Szegedy, C., Vanhoucke, V., Ioffe, S., Shlens, J., Wojna, Z., 2016. Rethinking the inception architecture for computer vision, in: Proceedings of the IEEE conference on computer vision and pattern recognition, pp. 2818–2826.

Tan, M., Le, Q., 2019. Efficientnet: Rethinking model scaling for convolutional neural networks, in: International Conference on Machine Learning, PMLR. pp. 6105–6114.

Wang, W., Lai, Q., Fu, H., Shen, J., Ling, H., Yang, R., 2021a. Salient object detection in the deep learning era: An in-depth survey. IEEE Transactions on Pattern Analysis and Machine Intelligence .

Wang, X., Jiang, X., Ding, H., Zhao, Y., Liu, J., 2021b. Knowledge-aware deep framework for collaborative skin lesion segmentation and melanoma recognition. Pattern Recognition , 108075.

Wei, C., Sohn, K., Mellina, C., Yuille, A., Yang, F., 2021. Crest: A class-rebalancing self-training framework for imbalanced semi-supervised learning, in: Proceedings of the IEEE/CVF Conference on Computer Vision and Pattern Recognition, pp. 10857–10866.

Weng, F., Meng, Y., Lu, F., Wang, Y., Wang, W., Xu, L., Cheng, D., Zhu, J., 2022. Differentiation of intestinal tuberculosis and crohn's disease through an explainable machine learning method. Scientific Reports 12, 1–12.

Weng, F., Zhang, H., Yang, C., 2021. Volatility forecasting of crude oil futures based on a genetic algorithm regularization online extreme learning machine with a forgetting factor: The role of news during the covid-19 pandemic. Resources Policy 73, 102148.

Zagoruyko, S., Komodakis, N., 2016. Wide residual networks. arXiv preprint arXiv:1605.07146 .

Zhao, Z.Q., Zheng, P., Xu, S.t., Wu, X., 2019. Object detection with deep learning: A review. IEEE transactions on neural networks and learning systems 30, 3212–3232.